\documentclass{imsart}
\RequirePackage[OT1]{fontenc}
\RequirePackage{amsthm,amsmath,natbib}
\RequirePackage[colorlinks,citecolor=blue,urlcolor=blue]{hyperref}

\usepackage{geometry}
\RequirePackage{amssymb,mathtools }
\usepackage{booktabs}
\usepackage[table]{xcolor}
\usepackage{algorithm,algpseudocode,float}
\usepackage{eurosym}
\usepackage{tikz}
\usetikzlibrary{fit,positioning,arrows,calc}
\tikzset{
  main/.style={circle, minimum size = 5mm, thick, draw =black!80, node distance = 10mm},
  connect/.style={-latex, thick},
  box/.style={rectangle, draw=black!100},
  edge/.style={->,> = latex'}
}
\setcitestyle{authoryear,open={(},close={)}}

\startlocaldefs
\numberwithin{equation}{section}
\theoremstyle{plain}

\theoremstyle{remark}

\endlocaldefs

\begin{document}

\begin{frontmatter}
\title{A new approach to learning in Dynamic Bayesian Networks (DBNs)}
\runtitle{Learning in Dynamic Bayesian Graphical Model}

\begin{aug}
  \author{\fnms{Eric} \snm{Benhamou}\thanksref{t1,t2} \corref{}  \ead[label=aisquare]{eric.benhamou@aisquareconnect.com}
  \ead[label=e1]{eric.benhamou@dauphine.eu}}
  \author{\fnms{Jamal} \snm{Atif}\thanksref{t2}
  \ead[label=e2]{jamal.atif@dauphine.fr}}
  \author{\fnms{Rida}  \snm{Laraki}\thanksref{t2}
  \ead[label=e3]{rida.laraki@dauphine.fr}}

\thankstext{t1}{A.I Square Connect}
\thankstext{t2}{Lamsade}

  \runauthor{E. Benhamou, J. Atif, R. Laraki}

  \affiliation{AI Square Connect and Lamsade  and INRIA}

  \address{\textsuperscript{*}A.I. SQUARE CONNECT, 35 Boulevard d'Inkermann 92200 Neuilly sur Seine, France, \\
          \printead{aisquare}}
          
  \address{$\dagger$LAMSADE (UMR CNRS 7243), Universit\'e Paris Dauphine, , 75016 Paris, France\\
        \printead{e1,e2,e3}}

\end{aug}

%


\begin{abstract}
In this paper, we revisit the parameter learning problem, namely the estimation of model parameters for Dynamic Bayesian Networks (DBNs). DBNs are directed graphical models of stochastic processes that encompasses and generalize Hidden Markov models (HMMs) and Linear Dynamical Systems (LDSs). Whenever we apply these models to economics and finance, we are forced to make some modeling assumptions about the state dynamics and the graph topology (the DBN structure). These assumptions may be incorrectly specified and contain some additional noise compared to reality. Trying to use a best fit approach through maximum likelihood estimation may miss this point and try to fit at any price these models on data. We present here a new methodology that takes a radical point of view and instead focus on the final efficiency of our model. Parameters are hence estimated in terms of their efficiency rather than their distributional fit to the data. The resulting optimization problem that consists in finding the optimal parameters is a hard problem. We rely on Covariance Matrix Adaptation Evolution Strategy (CMA-ES) method to tackle this issue. We apply this method to the seminal problem of trend detection in financial markets. We see on numerical results that the resulting parameters seem less error prone to over fitting than traditional moving average cross over trend detection and perform better. The method developed here for algorithmic trading is general. It can be applied to other real case applications whenever there is no physical law underlying our DBNs.
\end{abstract}

\begin{keyword}
\kwd{DBNs}
\kwd{CMA ES}
\kwd{trend detection}
\kwd{systematic trading}
\end{keyword}

\end{frontmatter}


\section{Introduction}
As stated in \cite{Jordan_2012} (first sentence of the preface), \textit{Graphical models are a marriage between probability theory and graph theory}. They are very powerful as they provide a condensed way to represent variables dependencies. The graphical representation allows not only compacting information. It also provides a powerful formalism for representing and reasoning under uncertainty. It can also represent knowledge about the dynamics of the variables and hence leads to Dynamic Bayesian Networks. 

DBNs extend the notion of Bayesian networks as they represents the evolution of the random variables as a function of a discrete sequence, usually a time steps sequence. Hence they are the natural tool to model discrete time chronological observation. 
As we progress over the sequence, the dynamic terms represented by the model parameters may change while the network architecture stays.The representation as a probabilistic graphical model structured in an acyclic oriented graph enables calculating efficiently conditional probabilities related to model variables. 

A typical example of a DBN would be, in medical diagnosis, to determine the probability for a patient to have or host a disease according to his symptoms. This system is made "dynamic" by incorporating the fact that the probability of being sick at time t also depends on past probabilities. Intuitively, this means that risk evolves over time. 

\subsection{DBN and DAG}
More formally, a Bayesian network is a directed acyclic graph (DAG) denoted by $G = (V, E)$, where $V$ is the set of nodes and $E$ the set of edges connecting the nodes. A conditional probability distribution is associated to each node $x$, and the factorized joint probability on the set of $V$ is given by

$$
\mathbb P (V) = \prod_ {x \in V} \mathbb P \big (x \, \big | \, \operatorname {\pi}_x \big) 
$$
where the parents of a node $x$ are denoted by $\operatorname {\pi}_{x}$. A dynamic Bayesian Network (DBN) is defined as a pair $(B_0, B_ {2d})$ where $B_0$ is a traditional Bayesian network representing the initial or \textit{a priori} distribution of random variables, that can be related to time 0 
and where $B_ {2d}$ is a dynamic two-step Bayesian network describing the transition from time $t-1$ to time $t$ with
the probability $\mathbb{P} (x_t \, \big | \, x_ {t-1}) $  for any node $x$ belonging to $V$, in a directed acyclic graph $G = (V, E)$. 
The joined probability for two sets of nodes $V_t$ and $V_{t-1}$ is given by

$$ 
\mathbb{P} (V_{t} \, \big | \, V_{t-1}) = \prod_ {x \in V, \operatorname {\pi}_{x} \in V} \mathbb P (x_ {t} \, \big | \, \operatorname {\pi}_{x_ {t}})) \, 
$$
The factorized joint probability law is computed by \textit{tracing} the sequence in the graph over the time sequence. If we denote by $T$ the total length of the path and by $\mathbf{P} (V_ {0})$ the joined probability of the initial network $B_0$, the probability to go from $V_0$ to $T$ is given by:

$$
\mathbb{P} (V_ {0: T}) = \mathbb{P} (V_ {0}) \times P (V_ {1: T}) = \prod_ {x \in V} \mathbb{P} (x_ {0} \, \big | \, \operatorname {\pi} (x_ {0})) \times \prod_ {t = 1} ^ T \prod_ {x \in V}\mathbb{P}  (x_ {t} \, \big | \, \operatorname {\pi} (x_ {t})) 
$$

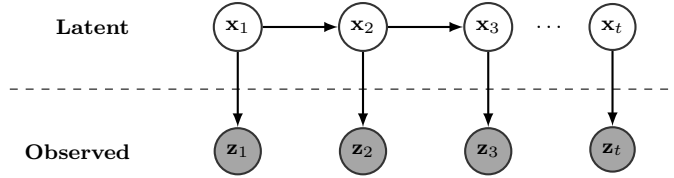
\begin{figure}[ht]
\centering

\begin{tikzpicture}
 \node[box,draw=white!100] (Latent) {\textbf{Latent}};
 \node[main] (L1) [right=of Latent] {$\mathbf{x}_1$};
 \node[main] (L2) [right=of L1] {$\mathbf{x}_2$};
 \node[main] (L3) [right=of L2] {$\mathbf{x}_3$};
 \node[main] (Lt) [right=of L3] {$\mathbf{x}_t$};
 \node[main,fill=black!35] (O1) [below=of L1] {$\mathbf{z}_1$};
 \node[main,fill=black!35] (O2) [below=of L2] {$\mathbf{z}_2$};
 \node[main,fill=black!35] (O3) [below=of L3] {$\mathbf{z}_3$};
 \node[main,fill=black!35] (Ot) [below=of Lt] {$\mathbf{z}_t$};
 \node[box,draw=white!100,left=of O1] (Observed) {\textbf{Observed}};

 \path (L1) edge [connect] (L2)
        (L2) edge [connect] (L3)
        (L3) -- node[auto=false]{\ldots} (Lt);

 \path (L1) edge [connect] (O1)
	(L2) edge [connect] (O2)
	(L3) edge [connect] (O3)
	(Lt) edge [connect] (Ot);

 \draw [dashed, shorten >=-1cm, shorten <=-1cm]
      ($(Latent)!0.5!(Observed)$) coordinate (a) -- ($(Lt)!(a)!(Ot)$);
\end{tikzpicture}
\caption{State Space model as a Bayesian Probabilistic Graphical model. Each vertical slice represents a time step. Nodes in white represent unobservable or latent variables  called the states and denoted by $\mathbf{x}_t$ while nodes in gray observable ones and are called the spaces and denoted by $\mathbf{z}_t$. Each arrow indicates that there is a relationship between the arrow originating node and the arrow targeting node. Dots indicate that there is many time steps. The central dot line is to emphasize the fundamental difference between latent and observed variables. This State Space model encompasses HMM and KF models} \label{HMM}
\end{figure}

\begin{figure}[ht]
\centering

\begin{tikzpicture}
 \node[box,draw=white!100] (Latent) {\textbf{Latent}};
 \node[main] (L0) [right=of Latent] {$\mathbf{x}_0$};
 \node[main] (L1) [right=of L0] {$\mathbf{x}_1$};
 \node[main] (L2) [right=of L1] {$\mathbf{x}_2$};
 \node[main] (L3) [right=of L2] {$\mathbf{x}_3$};
 \node[box,draw=white!100] (L4) [right=of L3] {};
 \node[main] (Lt) [right=of L4] {$\mathbf{x}_t$};

 \node[main] (L20) [below=of L0 ] {$\mathbf{y}_0$};
 \node[main] (L21) [below=of L1 ] {$\mathbf{y}_1$};
 \node[main] (L22) [below=of L2] {$\mathbf{y}_2$};
 \node[main] (L23) [below=of L3] {$\mathbf{y}_3$};
 \node[box,draw=white!100] (L24) [right=of L23] {};
 \node[main] (L2t) [below=of Lt] {$\mathbf{y}_t$};

 \node[main,fill=black!35] (O1) [below=of L21] {$\mathbf{z}_1$};
 \node[main,fill=black!35] (O2) [below=of L22] {$\mathbf{z}_2$};
 \node[main,fill=black!35] (O3) [below=of L23] {$\mathbf{z}_3$};
 \node[box,draw=white!100] (O4) [below=of L24] {};
 \node[main,fill=black!35] (Ot) [below=of L2t] {$\mathbf{z}_t$};
 \node[box,draw=white!100,below=90pt] (Observed) {\textbf{Observed}};

 \path (L0) edge [connect] (L1)
	 (L1) edge [connect] (L2)
        (L2) edge [connect] (L3)
        (L3) -- node{\ldots} (L4)
        (L4) edge [connect] (Lt);

 \path (L20) edge [connect] (L21)
	(L21) edge [connect] (L22)
        (L22) edge [connect] (L23)
        (L23) -- node{\ldots} (L24)
        (L24) edge [connect] (L2t);

 \path (L21) edge [connect] (O1)
	(L22) edge [connect] (O2)
	(L23) edge [connect] (O3)
	(L2t) edge [connect] (Ot);

 \path (L1) edge [bend right,connect] (O1)
	(L2) edge [bend right, connect] (O2)
	(L3) edge [bend right, connect] (O3)
	(Lt) edge [bend right, connect] (Ot);

 \path (L0) edge [connect] (L21)
	 (L1) edge [connect] (L22)
        (L2) edge [connect] (L23)
        (L4) edge [connect] (L2t);

 \draw [dashed, shorten >=-0.5cm, shorten <=-4cm]
     ($(O1)+(0,0.8)$) -- ($(Ot)+(0,0.8)$);
\end{tikzpicture}
\caption{Example of another dynamic Bayesian network combining Kalman filter (KF) model (and echo neural networks (ESN). This is another example of a multi-input several multi-outputs (MISMO) forecasting model. It is used frequently in time series forecast (see for instance \cite{Xiao_2017})} \label{DBN2}
\end{figure}
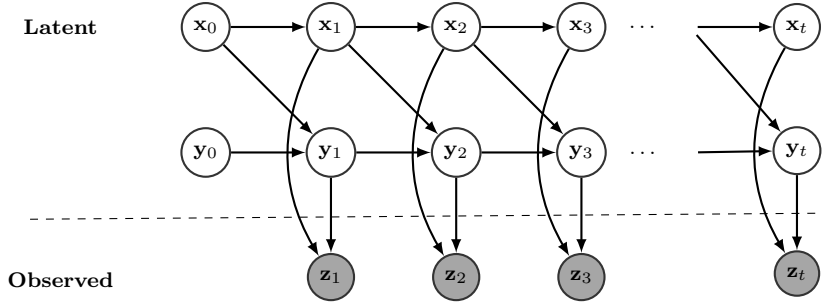

A dynamic Bayesian network thus respects the \textit{Markov property}, which expresses that conditional distributions at time $t$ depend only on the state at time $t-1$. Dynamic Bayesian networks generalize probabilistic models such as Hidden Markov Model (HMM), and Kalman filter (KF). 
Apart from the mainstream Kalman filter and HMM models whose DAG is given by \ref{HMM}, more complex DBN can include multi input network with connection between observable and previous latent variables as provided by \ref{DBN1}. Another example is the combination of Kalman Filtering (KF) model and echo neural network (ESN) as provided by figure \ref{DBN2}. 

\begin{figure}[ht]
\centering

\begin{tikzpicture}
 \node[box,draw=white!100] (Latent) {\textbf{Latent}};
 \node[main] (L1) [right=of Latent] {$\mathbf{x}_1$};
 \node[main] (L2) [right=of L1] {$\mathbf{x}_2$};
 \node[main] (L3) [right=of L2] {$\mathbf{x}_3$};
 \node[box,draw=white!100] (L4) [right=of L3] {};
 \node[box,draw=white!100] (L5) [right=of L4] {\ldots};
 \node[main] (Lt) [right=of L5] {$\mathbf{x}_t$};

 \node[main] (L21) [below right=of L1 ] {$\mathbf{y}_1$};
 \node[main] (L22) [below right=of L2] {$\mathbf{y}_2$};
 \node[main] (L23) [below right=of L3] {$\mathbf{y}_3$};
 \node[main] (L2t) [below right=of Lt] {$\mathbf{y}_t$};

 \node[main,fill=black!35] (O1) [below right=of L21] {$\mathbf{z}_1$};
 \node[main,fill=black!35] (O2) [below right=of L22] {$\mathbf{z}_2$};
 \node[main,fill=black!35] (O3) [below right=of L23] {$\mathbf{z}_3$};
 \node[main,fill=black!35] (Ot) [below right=of L2t] {$\mathbf{z}_t$};
 \node[main,fill=black!35,scale=0.7] (Otm1) [left=of Ot] {$\mathbf{z}_{\small{t-1}}$};
 \node[box,draw=white!100,below=60pt] (Observed) {\textbf{Observed}};

 \path (L1) edge [connect] (L2)
        (L2) edge [connect] (L3)
        (L3) -- node{\ldots} (Lt);

 \path (L21) edge [connect] (L22)
        (L22) edge [connect] (L23)
        (L23) -- node{\ldots} (L2t);

 \path (L21) edge [connect] (O1)
	(L22) edge [connect] (O2)
	(L23) edge [connect] (O3)
	(L2t) edge [connect] (Ot);

 \path (L1) edge [bend right, connect] (O1)
	(L2) edge [bend right, connect] (O2)
	(L3) edge [bend right, connect] (O3)
	(Lt) edge [bend right, connect] (Ot);

 \path (O1) edge [connect] (L22)
	(O2) edge [connect] (L23)
	(Otm1) edge [connect] (L2t);

 \draw [dashed, shorten >=-2cm, shorten <=-4cm]
      ($(L21)-(0,0.6)$) -- ($(L2t)-(0,0.6)$);
\end{tikzpicture}
\caption{Example of a dynamic Bayesian network where we have some variables that are observed and some that are latent (non observable). 
We can see the DAG structure of the network and the Markovian property. This graphical model is more complexed than the traditional state space model as it includes in each time step multi input variables as well as connection between past observable variables and latent variables} \label{DBN1}
\end{figure}
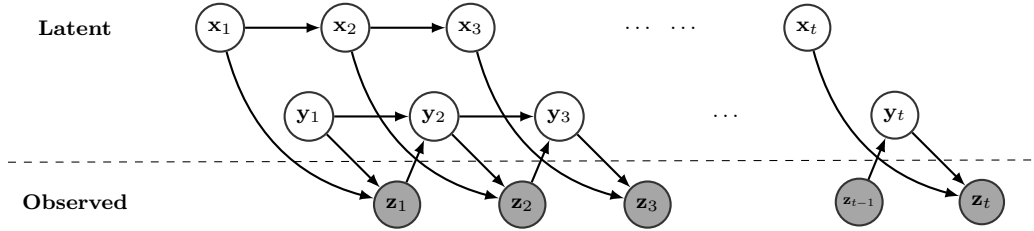

\subsection{EM method and its drawback}
A majority of DBNs exploit latent variables to make the model more powerful in terms of explanation power. 
By defining a joint distribution over visible and latent variables, the corresponding distribution of the observed variables is obtained
by marginalization. This has the nice property to express relatively complex distributions in terms of more tractable joint distributions over the expanded
variable space. One well-known example of a hidden variable model is the mixture distribution in which the hidden variable is the discrete component label that provides the corresponding distribution for the observable variable. The static version leads to Gaussian mixture model and more generally the factor analysis model while the dynamic version leads to HMM and in continuous time space to Kalman filter model. However, this does not solve the issue of learning the model parameters. The typical approach is to use the Expectation Maximization (EM) approach to find the gradient coordinate ascent of the Kullback Leibler divergence. The EM algorithm was initially developed for mixture models in particular Gaussian mixtures but also other natural laws from the exponential family such as Poisson, binomial, multinomial and exponential distributions as early as in \cite{Hartley_1958}. It was only once the link between latent variable and Kalman filter models was made that it became obvious that this could also be applied to Kalman and extended Kalman filter (see \cite{Cappe_2010} or \cite{Einicke_2010}). The EM method is sofar the state of the art method for learning DBNs as it provides an efficient way to find model parameters in a fraction of seconds (see for instance \cite{Neal_1999}, but also \cite{Pfeifer_2018}, \cite{Li_2017}, \cite{Robin_2017}, \cite{Levine_2018}).

However, we argue that although this is a nice method, it misses the point that the DBN is an imprecise model of the reality especially when tackling problem like time series forecasting. 
In particular, whenever we apply DBNs to economics and finance, we are forced to make some modeling assumptions about the state dynamics and the graph topology (the DBN structure). These assumptions may be incorrectly specified and contain some additional noise compared to reality. Trying to use a best fit approach through maximum likelihood estimation and Kullback Leibler divergence optimization may miss this point and try to fit at any price these models on data. We present here a new methodology that takes a radical point of view and instead focus on the final efficiency of our model. Parameters are hence estimated in terms of their efficiency rather than their distributional fit to the data. Our approach relies on Information Geometry optimization and find a local optimum for our final cost function.
Our key findings are the following:
\begin{itemize}
\item it is possible to directly optimize the cost function with an Evolution Strategies (ES) and in particular the CMA-ES method.
\item this approach is a good alternative to the EM approach as it does not fit at all cost the distribution of our network to reality but rather look at model efficiency measured by model cost.
\item numerical results shows that the overfitting issue of this approach due to local minimum is less than the EM approach as it incorporates somehow that the model dynamics is incorrectly specified and too simple.
\end{itemize}

The rest of the paper is organize as follows. Section \ref{Settings} presents the overall framework and the resulting optimization problem. Section \ref{Properties} provides some theoretical arguments that favor Evolution Strategies based on Information Geometry Optimization (IGO). Section \ref{NumericalResults} provides an example in finance of such a method. The method outperforms traditional trend following method by far. We finally conclude about possible extensions of this method and further experiments.

\section{Settings} \label{Settings}
Suppose we have determined an architecture for our network. This may be any of the networks provided by figures \ref{HMM}, \ref{DBN1}, \ref{DBN2}, or even something different. This model is used for some specific goal. In our case, it will be used to be able to forecast some times series. But this is not our final objective! We are interested in using this forecast to perform a specific action. In the case of a financial markets algorithmic trading strategy, we will use the forecast to make an informed decision and decide whether we should buy or sell a given financial asset. To make things simple, we will assume that when we take our decision, we have in mind a pre-determined strategy. We could imagine a dynamic approach where as time passes we change our objective once we have executed our entry order. For the sake of simplicity, we will assume that whenever we issue an order to trade, we have determined a profit target and a stop loss level for our trade. Although we may have simplified a little bit our setting, this assumption is quite realistic and done by many practitioners as stated in various papers (\cite{Mauricio_2010}, \cite{Graziano_2014}, \cite{Stanley_2017}, or \cite{Vezeris_2018}). The profit target ensures that the strategy locks in real money the profit realized and is materialized by a limit order. The stop loss that is physically generated by a stop order safeguards the overall risk by limiting losses whenever the market backfires and contradicts the presumed pattern. 

\begin{figure}
       \includegraphics[width=12cm]{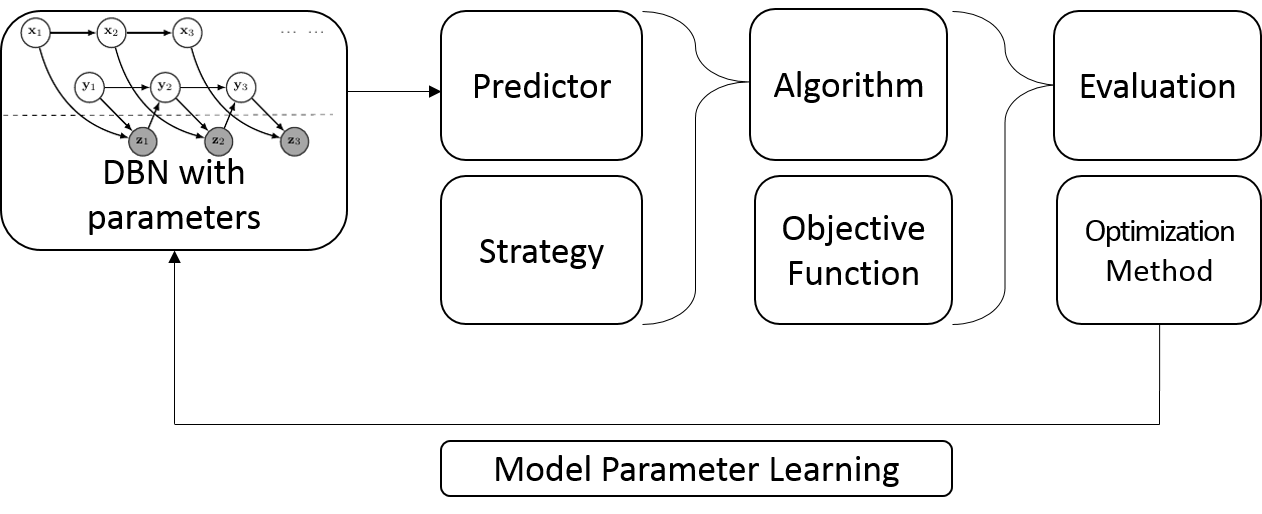}
	\caption{Learnig process for our DBN. Once an architecture has been decided, we combine this with a strategy to create an systematic algorithm. We select an objective function that provides an Evaluation function. We use an optimization method to find the best parameters according to our evaluation function. We monitor overall performance of the trading strategy on a separate test set to validate scarce overfitting.}\label{challenge}
	\centering
\end{figure}

Our final goal at the end of the day is to generate a profitable strategy that does not suffer from overfitting. This is really what matters for us. Hence the EM approach that tackles the overall fit of our model to reality may be inappropriate. This may look quite simple but there are here some real complexity. The overall objective function is often a complex function. In our case, we are interested in maximizing the overall Sharpe ratio of our trading strategy over time and check that the observed performance on the training set does not vanish on the test set. The optimization problem that we are facing is not a simple one as we want to maximize our objective function with respect to our DBN model. The overall goal is summarized in figure \ref{challenge}

\subsection{CMA-ES estimation}
One of the key point in the selected architecture is the optimization method. In our case, the decision strategy works as follows. If the DBN predicts that the next value of the time series is higher than the last know one plus an offset, we decide to buy the financial asset. Similarly if the DBN predicts that the next value is lower with a sell action. Whenever we issue an order to trade, we also set a profit target and a stop loss in percentage of the last know value. We stay in position until either the trade reaches the profit target and we exit with a profit or it touches the stop loss level and we exit with a loss. The objective function to measure the overall performance of the model is the so called Sharpe ratio. This is a usual performance metric that was established in \cite{Sharpe_1966}. It is defined in our case as we target absolute performance as trades gains and losses average over the standard deviation. This number is easy to derive and intuitive to understand as it computes the ratio of the excess return over the strategy standard deviation.

Clearly our objective function is non convex and quite complex to evaluate. It has some discontinuities whenever we reaches critical values. Hence, we rely on CMA-ES algorithm to optimize this evaluation function. CMA ES name stands for covariance matrix adaptation evolution strategy. As it points out, it is an evolution strategy optimization method, meaning that it is a derivative free method that can accommodate non convex optimization problem. The terminology covariance matrix alludes to the fact that the exploration of new points is based on a multinomial distribution whose covariance matrix is progressively determined at each iteration. Hence the covariance matrix adapts in a sense to the sampling space, contracts in dimension that are useless and expands in dimension where natural gradient is steep. This algorithm has led to a large number of papers and articles and we refer to \cite{Hansen_2018}, \cite{Ollivier_2017}, \cite{Auger_2016}, \cite{Auger_2015}, \cite{Hansen_2014}, \cite{Auger_2012}, \cite{Hansen_2011}, \cite{Auger_2009}, \cite{Igel_2007}, \cite{Auger_2004} to cite a few of the numerous articles around CMA-ES. We also refer the reader to the excellent wikipedia page \cite{wiki:CMAES}.

CMA-ES relies on two main principles in the exploration of admissible solution for our optimization problem.
First, it relies on a multi variate normal distribution as this is the maximum entropy distribution given the first two moments.
The mean of the multi variate distribution is updated at each step in order to maximize the likelihood of finding a successful candidate. The second moment, the covariance matrix of the distribution is also updated at each step to increase the likelihood of successful search steps. These updates can be interpreted as a natural gradient descent \cite{Ollivier_2017}. 

Second, we retain two paths of the successive distribution mean, called search or evolution paths. The underlying idea is keep significant information about the correlation between consecutive steps. If consecutive steps are taken in a similar direction, the evolution paths become long. The evolution paths are exploited in two ways. We use the first path is to compute the covariance matrix to increase variance in favorable directions and hence increase convergence speed. The second path is used to control step size and to make consecutive movements of the distribution mean orthogonal in expectation. The goal of this step-size control is to prevent premature convergence yet obtaining fast convergence to a local optimum.

From a practical point of view, we assume that we have a general cost function that depends on our Bayesian graphical model denoted by $\Phi( \theta)$ where $\theta$ are the parameters of our Kalman filter. Our cost function is the Sharpe ratio corresponding to a generic trend detection strategy whose signal is generated by our Bayesian graphical model that is underneath a Kalman filter. This approach is more developed in a companion paper \cite{Benhamou_2018_CMAES} but we will give here the general idea. Instead of computing the parameter of our Bayesian graphical model using the EM approach, we would like to find the parameters $\theta_{\max}$ that maximize our cost function $\Phi(\theta)$. Because our cost function is to enter a long trade with a predetermined target level and a given stop loss whenever our Bayesian graphical model anticipates a price risen and similarly to enter a short trade whenever our prediction based on Bayesian graphical model is a downside movement, our trading strategy is not convex neither smooth. It is a full binary function and generates spike whenever there is a trade. Moreover, our final criterium is to use the Sharpe ratio of the resulting trading strategy to compare the efficiency of our parameters. This is way too complicated for traditional optimization method, and we need to rely on Black box optimization techniques like CMA-ES. 

\section{Properties} \label{Properties}
\subsection{Maximum entropy}
One of the theoretical justification of the CMA-ES algorithm  is that the multivariate normal distribution for sampling new candidate solutions is the maximum entropy probability distribution over $\mathbb{R}^n$, which translates that it is the  sample distribution with the minimal amount of prior information. 

\subsection{Maximum-likelihood updates}
The update for the mean and covariance matrix are built such as to maximize the empirical likelihood. 
This is in a sense quite similar to the Expectation maximization method as the updates are done first by computing the expectation and then taking the maximum, which leads to
\begin{equation}
m_{k+1} = \arg\max_{m} \sum_{i=1}^\mu w_i \log p_\mathcal{N}(x_{i:\lambda} | m) 
\end{equation}

where the log-likelihood of $x$ assuming a multivariate normal distribution with mean $m$ and any positive definite covariance matrix $C$ is given by
\begin{equation}
\log p_\mathcal{N}(x) = 
   - \frac{1}{2} \log\det(2\pi C) - \frac{1}{2} (x-m)^T C^{-1} (x-m) 
\end{equation}
   
\subsection{Natural gradient descent}
Last but not least, an interesting feature that provides theoretical justification for the CMA-ES algorithm and that was found by various authors (\cite{Akimoto_2010,Akimoto_2012}, \cite{Glasmachers_2010} and also the seminal work of \cite{Ollivier_2017}) is that the parameters update realizes a natural gradient descent in the space of the sample distributions

\section{Numerical Results} \label{NumericalResults}
In order to test the efficiency of the CMA Es method for Learning parameters in DBNs, we look at the following trend following algorithm based on the following DBN network where we enter a long trade if the prediction of our dynamic Bayesian network forecast is above the close of the previous day and a short trade if the prediction is below the close of the previous day. For each comparison, we add an offset $\mu$ to avoid triggering false alarm signals. We set for each trade a pre-determined profit and stop loss target in ticks. These parameters are optimized in order to provide the best sharpe ratio over the train period together with the DBN parameters given by

We take the following HMM model for our DBN (for more details, see \cite{Benhamou_Kalman})

\begin{align}
   \mathbf{x}_{t+1} & = \mathbf{\Phi}    \mathbf{x}_t + \mathbf{c}_t +  \mathbf{w}_t \label{model_eq1} \\
   \mathbf{z}_t        & = \mathbf{H} \mathbf{x}_t +  \mathbf{v}_t \label{model_eq2} 
\end{align}   

The noise process $\mathbf{w}_t$ is assumed to follow a multi dimensional normal distribution with zero mean and covariance matrix given by $\mathbf{Q}_t$: $\mathbf{w}_t \sim \mathcal{N}\left(0, \mathbf{Q}_t\right)$.

We also assume that the observation noise $\mathbf{v}_t$ follows a multi dimensional normal distribution with zero mean and covariance matrix given by $\mathbf{R}_t$: $\mathbf{v}_t \sim \mathcal{N}\left(0, \mathbf{R}_t\right)$. In addition, the initial state, and noise vectors at each step ${\mathbf{x}_0, \mathbf{w}_1, \ldots, \mathbf{w}_t, \mathbf{v}_1, \ldots, \mathbf{v}_t}$ are assumed to be all mutually independent. We also denote by $\mathbf{P}_t=\operatorname{Cov}(\mathbf{x}_{t})$ the covariance matrix of $\mathbf{x}_{t}$. We assume the following parameters:

\begin{eqnarray}
& & \mathbf{\Phi}(x) = \left[ {\begin{array}{cc} \!\!  p_1 & p_2 \! \!  \\ \!\!   0 &  p_3 \!\!   \end{array} } \right]  \quad \mathbf{H} = \left[ {\begin{array}{c} \!\! p_4 \!\!  \\  \!\!  p_5 \! \! \end{array} } \right]  \quad \mathbf{Q}_{t=0}= \left[ {\begin{array}{l}  \!  p_6^2  \; p_6p_7 \!  \\ \!  p_7 p_6 \;   p_8^2  \!  \end{array} } \right]  \quad \mathbf{R}_{t=0} = \left[ {\begin{array}{c} \! p_9 \!  \end{array} } \right]   \label{model_eq3}  \\
& & \mathbf{P}_{t=0}  = \left[ {\begin{array}{l}  \!\!  p_{10}  \; 0 \!\!  \\ \!\! 0 \;   p_{11}  \! \! \end{array} } \right]  \qquad 
\mathbf{c}_t = \left[ {\begin{array}{c} \!\!  p_{12} ( p_{13} - K_t  )\! \!  \\ \!\!  p_{14}  ( p_{15} - K_t  )\!\!  \end{array} } \label{model_eq4}  \right]  
\end{eqnarray}

The pseudo code of our algorithm is listed below

\begin{algorithm}[H]
\caption{Kalman filter Trend following algorithm}
	\begin{algorithmic} 
	\State \textbf{Initialize common trade details}
	\State SetProfitTarget( target)							\Comment{fixed profit target in ticks}
	\State SetStopLoss( stop\_loss )							\Comment{fixed stop loss in ticks}
	\\
	\While{ Not In Position}											\Comment{look for new trade}
		\If{ DBN( $p_1, \ldots, p_n$).Predict[0] $\ge$ Close[0] + $\mu$} 	\Comment{up trend signal}
			\State EnterLong()										\Comment{market order for the open}
		\ElsIf{ DBN( $p_1, \ldots, p_n$).Predict[0] $\le$ Close[0] + $\mu$} 	\Comment{down trend signal}
			\State EnterShort()										\Comment{market order for the open}
		\EndIf
	\EndWhile
	\end{algorithmic}
\end{algorithm}

Our resulting algorithm depends on the following parameters $p_1, \ldots, p_n$ the Kalman filter algorithm, the profit target, the stop loss and the signal offset $\mu$. We could estimate the Kalman filter parameters with the EM procedure, then optimize the profit target, the stop loss and the signal offset $\mu$. However, if by any chance the dynamics of the Kalman filter is incorrectly specified, the noise generated by this wrong specification will only be factored in the three parameters:  the profit target, the stop loss and the signal offset $\mu$. We prefer to do a combined optimization of all the parameters. We use daily data of the S\&P 500 index futures (whose CQG code is \textit{EP}) from 01Jan2017 to 01Jan2018. We train our model on the first 6 months and test it on the next six months. Deliberately, our algorithm is unsophisticated to keep thing simple and concentrate on the parameter estimation method. The overall idea is for a given set of parameter to compute the resulting sharpe ratio over the train period and find the optimal parameters combination. For a model given by equations \eqref{model_eq1} and \eqref{model_eq2}  and parameters specified in \eqref{model_eq3}  and \eqref{model_eq3}, the optimization encompasses 18 parameters: $p_1, \ldots, p_{15}$, the profit target, the stop loss and the signal offset $\mu$, making it non trivial. We use the CMA-ES algorithm to find the optimal solution. In our optimization, we add some penalty condition to force non meaningful Kalman filter parameters to be zero, namely, we add a L1 penalty on this parameters.

Results are given below

\hspace{-1.5cm}
\begin{minipage}{1.2\textwidth}

\begin{table}[H]
  \centering
  \caption{Optimal parameters}
\resizebox{\textwidth}{!}{
    \begin{tabular}{|c|r|r|r|r|r|r|r|r|r|r|r|r|r|r|r|r|r|r|}
    \toprule
    \newline{}\newline{}\newline{}\newline{}\newline{}Parameters & \multicolumn{1}{l|}{$p_{01}$} & \multicolumn{1}{l|}{$p_{02}$} & \multicolumn{1}{l|}{$p_{03}$} & \multicolumn{1}{l|}{$p_{04}$} & \multicolumn{1}{l|}{$p_{05}$} & \multicolumn{1}{l|}{$p_{06}$} & \multicolumn{1}{l|}{$p_{07}$} & \multicolumn{1}{l|}{$p_{08}$} & \multicolumn{1}{l|}{$p_{09}$} & \multicolumn{1}{l|}{$p_{10}$} & \multicolumn{1}{l|}{$p_{11}$} & \multicolumn{1}{l|}{$p_{12}$} & \multicolumn{1}{l|}{$p_{13}$} & \multicolumn{1}{l|}{$p_{14}$} & \multicolumn{1}{l|}{$p_{15}$} & \multicolumn{1}{l|}{offset} & \multicolumn{1}{l|}{stop} & \multicolumn{1}{l|}{target} \\
    \midrule
    Value & 24.8  & 0     & 11.8  & 46.2  & 77.5  & 67    & 100   & 0     & 0     & 0     & 0     & 100   & 0     & 0     & 0     & 5     & 80    & 150 \\
    \bottomrule
    \end{tabular}}%
  \label{tab:param}%
\end{table}%

\begin{table}[H]
  \centering
  \caption{Train test statistics 1/4}
\resizebox{\textwidth}{!}{
    \begin{tabular}{|c|c|c|c|c|c|c|c|c|}
    \toprule
    Performance & Net Profit & Gross Profit & Gross Loss & \# of Trades & \# of Contracts & Avg. Trade & Tot. Net Profit (\%) & Ann. Net Profit (\%) \\
    \midrule
    Train & \textcolor[rgb]{ 0,  .502,  0}{5,086 \euro} & \textcolor[rgb]{ 0,  .502,  0}{11,845 \euro} & \textcolor[rgb]{ 1,  0,  0}{-6,759 \euro} & 15    & 15    & \textcolor[rgb]{ 0,  .502,  0}{339.05 \euro} & \textcolor[rgb]{ 0,  .502,  0}{5.09\%} & \textcolor[rgb]{ 0,  .502,  0}{10.59\%} \\
    \midrule
    Test  & \textcolor[rgb]{ 0,  .502,  0}{4,266 \euro} & \textcolor[rgb]{ 0,  .502,  0}{11,122 \euro} & \textcolor[rgb]{ 1,  0,  0}{-6,857 \euro} & 15    & 15    & \textcolor[rgb]{ 0,  .502,  0}{284.38 \euro} & \textcolor[rgb]{ 0,  .502,  0}{4.27\%} & \textcolor[rgb]{ 0,  .502,  0}{8.69\%} \\
    \bottomrule
    \end{tabular}}
  \label{tab:stat1}%
\end{table}%

\begin{table}[H]
  \centering
  \caption{Train test statistics 2/4}
\resizebox{\textwidth}{!}{
    \begin{tabular}{|c|c|c|c|c|c|c|c|c|}
    \toprule
    Performance & Vol   & Sharpe Ratio & Trades per Day & Avg. Time in Market & Max. Drawdown & Recovery Factor & Daily Ann. Vol & Monthly Ann. Vol \\
    \midrule
    Train & \textcolor[rgb]{ 0,  .502,  0}{6.54\%} & \textcolor[rgb]{ 0,  .502,  0}{1.62} & 0.10  & 8d14h & \textcolor[rgb]{ 1,  0,  0}{-2,941 \euro} & \textcolor[rgb]{ 0,  .502,  0}{3.510} & \textcolor[rgb]{ 0,  .502,  0}{6.54\%} & \textcolor[rgb]{ 0,  .502,  0}{5.72\%} \\
    \midrule
    Test  & \textcolor[rgb]{ 0,  .502,  0}{6.20\%} & \textcolor[rgb]{ 0,  .502,  0}{1.40} & 0.10  & 8d19h & \textcolor[rgb]{ 1,  0,  0}{-1,721 \euro} & \textcolor[rgb]{ 0,  .502,  0}{4.948} & \textcolor[rgb]{ 0,  .502,  0}{6.20\%} & \textcolor[rgb]{ 0,  .502,  0}{5.32\%} \\
    \bottomrule
    \end{tabular}}%
  \label{tab:stat2}%
\end{table}%

\begin{table}[H]
  \centering
  \caption{Train test statistics 3/4}
\resizebox{\textwidth}{!}{
    \begin{tabular}{|c|c|c|c|c|c|c|c|c|}
    \toprule
    Performance & Daily Sharpe Ratio & Daily Sortino Ratio & Commission & Percent Profitable & Profit Factor & \# of Winning Trades & Avg. Winning Trade & Max. conseq. Winners \\
    \midrule
    Train & \textcolor[rgb]{ 0,  .502,  0}{1.62} & \textcolor[rgb]{ 0,  .502,  0}{2.35} & \textcolor[rgb]{ 1,  0,  0}{49 \euro} & 46.67\% & 1.75 \euro & 7     & \textcolor[rgb]{ 0,  .502,  0}{1,692.09 \euro} & 3 \\
    \midrule
    Test  & \textcolor[rgb]{ 0,  .502,  0}{1.40} & \textcolor[rgb]{ 0,  .502,  0}{2.05} & \textcolor[rgb]{ 1,  0,  0}{46 \euro} & 46.67\% & 1.62 \euro & 7     & \textcolor[rgb]{ 0,  .502,  0}{1,588.92 \euro} & 2 \\
    \bottomrule
    \end{tabular}}%
  \label{tab:stat3}%
\end{table}%

\begin{table}[H]
  \centering
  \caption{Train test statistics 4/4}
\resizebox{\textwidth}{!}{
    \begin{tabular}{|c|c|c|c|c|c|c|c|c|}
    \toprule
    Performance & Largest Winning Trade & \# of Losing Trades & Avg. Losing Trade & Max. conseq. Losers & Largest Losing Trade & Avg. Win/Avg. Loss & Avg. Bars in Trade & Time to Recover \\
    \midrule
    Train & \textcolor[rgb]{ 0,  .502,  0}{1,776.11 \euro} & 8     & \textcolor[rgb]{ 1,  0,  0}{-844.85 \euro} & 3     & \textcolor[rgb]{ 1,  0,  0}{-1,011.82 \euro} & 2.00  & 6.1   & 77.00 days \\
    \midrule
    Test  & \textcolor[rgb]{ 0,  .502,  0}{1,609.32 \euro} & 8     & \textcolor[rgb]{ 1,  0,  0}{-857.1 \euro} & 2     & \textcolor[rgb]{ 1,  0,  0}{-860.26 \euro} & 1.85  & 6.2   & 70.00 days \\
    \bottomrule
    \end{tabular}}%
  \label{tab:stat4}%
\end{table}%

\end{minipage}

\newpage

\begin{figure}
\centering
\includegraphics[width=13cm]{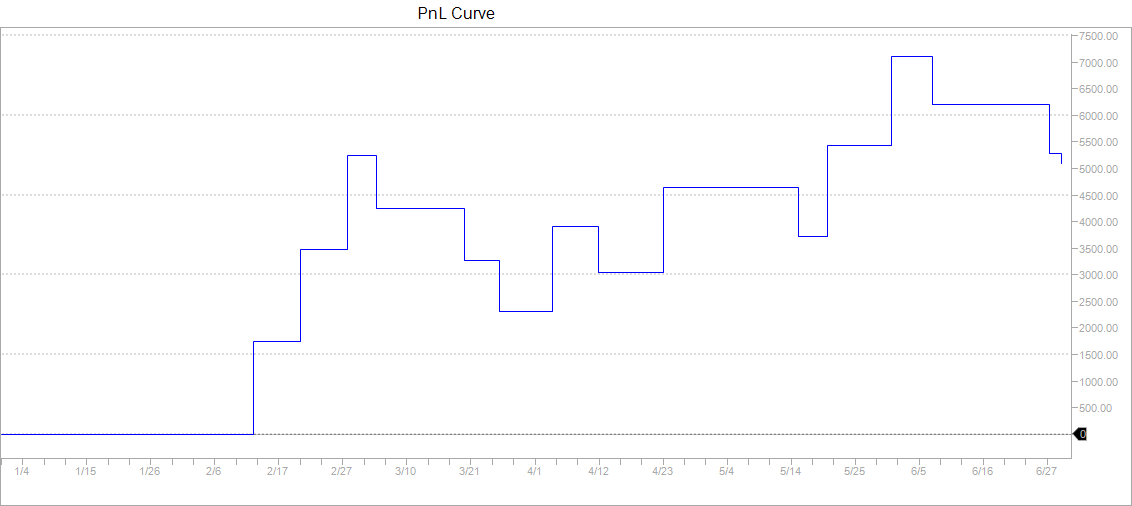}
\caption{Kalman filter algorithm on train data set}
\label{fig:train}
\end{figure}

\begin{figure}
\centering
\includegraphics[width=13cm]{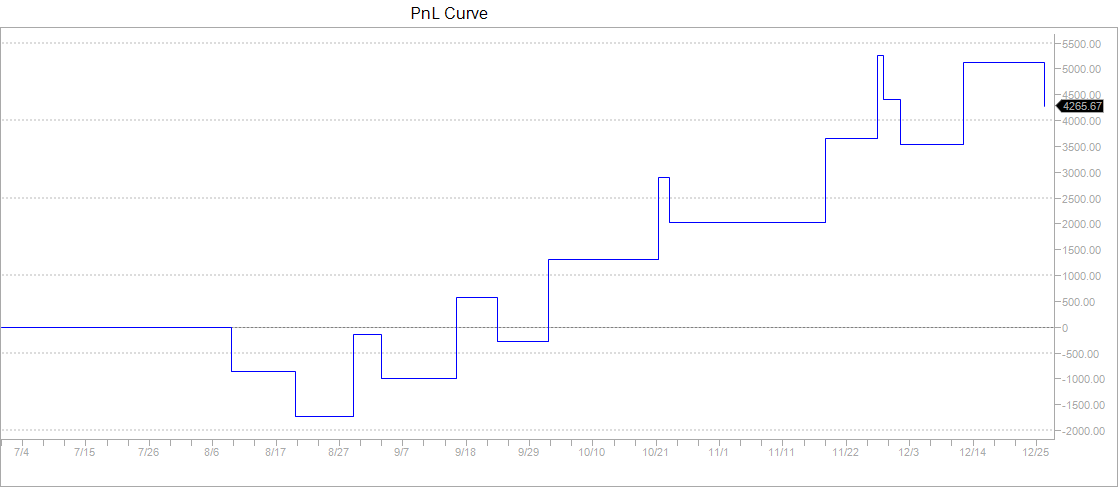}
\caption{Kalman filter algorithm on test data set}
\label{fig:test}
\end{figure}

We compare our algorithm with a traditional moving average crossover algorithm to test the efficiency of Kalman filter for trend detection. The moving average cross over algorithm generates a buy signal when the fast moving average crosses over the long moving average and a sell signal when the former crosses below the latter. A $d$ period moving average is defined as the arithmetic average of the daily close over a $d$ period, denoted by $\mathrm{SMA}(d)$. Our algorithm is given by the following pseudo code

\begin{algorithm}[H]
\caption{Moving Average Trend following algorithm}
	\begin{algorithmic} 
	\State \textbf{Initialize common trade details}
	\State SetProfitTarget( target)							\Comment{fixed profit target in ticks}
	\State SetStopLoss( stop\_loss )							\Comment{fixed stop loss in ticks}
	\\
	\While{ Not In Position}											\Comment{look for new trade}
		\If{ $\mathrm{SMA}\text{(Short)}[0] > \mathrm{SMA}\text{(Long)}[0] + \text{offset}$ } 	\Comment{up trend signal}
			\State EnterLong()										\Comment{market order for the open}
		\ElsIf{  $\mathrm{SMA}\text{(Short)}[0] < \mathrm{SMA}\text{(Long)}[0] + \text{offset}$ } 	\Comment{down trend signal}
			\State EnterShort()										\Comment{market order for the open}
		\EndIf
	\EndWhile
	\end{algorithmic}
\end{algorithm}

We can now compare moving average cross over versus Kalman filter algorith. The table \ref{tab:mavskf} compares the two algorithms. We can see that on the train period, the two algorithms have similar performances : $ 5,260$ vs $5,086$. However on the test period, moving average performs very badly with a net profit of $935$ versus $4,266$ for the bayesian graphical model (the kalman filter) algorithm.

\hspace{-1.5cm}
\begin{minipage}{1.2\textwidth}

\begin{table}[H]
  \centering
  \caption{Moving average cross over versus Kalman filter}
\resizebox{\textwidth}{!}{
    \begin{tabular}{|c|c|c|c|c|c|c|c|c|}
    \toprule
    Algo  & Total Net Profit & Recovery Factor & Profit Factor & Max. Drawdown & Sharpe Ratio & Total \# of Trades & Percent Profitable & Train: Total Net Profit \\
    \midrule
    \textcolor[rgb]{ .125,  .122,  .208}{MA Cross over} & \textcolor[rgb]{ 0,  .502,  0}{935 \euro} &                    0.32  &             1.13  & \textcolor[rgb]{ 1,  0,  0}{-\euro 2,889} &              0.41  &                         26  &                      0.54  &                     5,260.00  \\
    \midrule
    \textcolor[rgb]{ .125,  .122,  .208}{Kalman filter} & \textcolor[rgb]{ 0,  .502,  0}{4,266 \euro} &                    2.48  &             1.62  & \textcolor[rgb]{ 1,  0,  0}{-\euro 1,721} &              1.40  &                         30  &                      0.47  &                     5,085.79  \\
    \bottomrule
    \end{tabular}}%
  \label{tab:mavskf}%
\end{table}
\end{minipage}

\begin{figure}
\centering
\includegraphics[width=13cm]{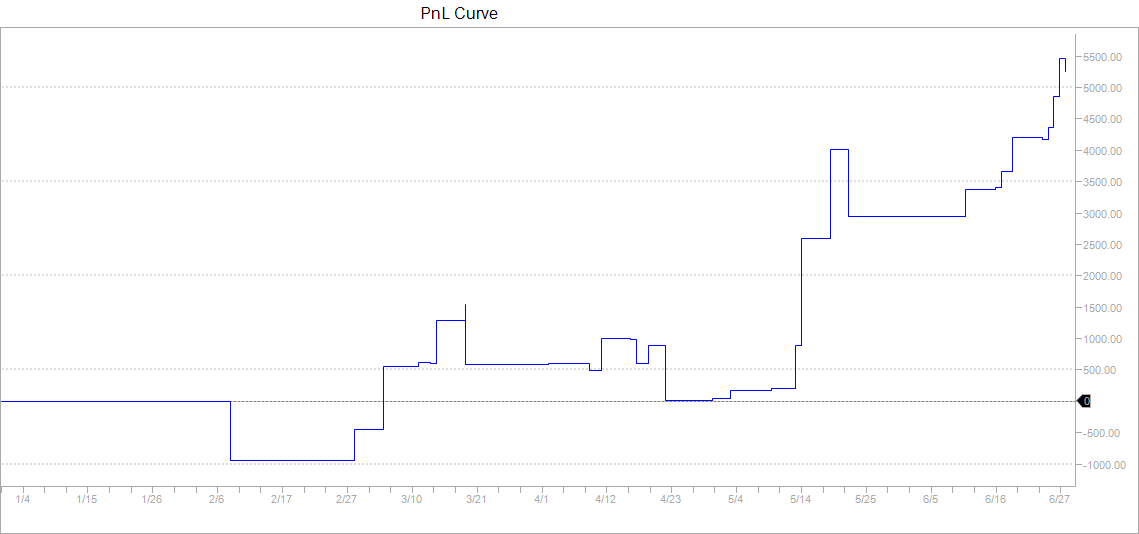}
\caption{Moving Average Crossover algorithm on train data set}
\label{fig:ma_train}
\end{figure}

\newpage
\begin{figure}
\centering
\includegraphics[width=13cm]{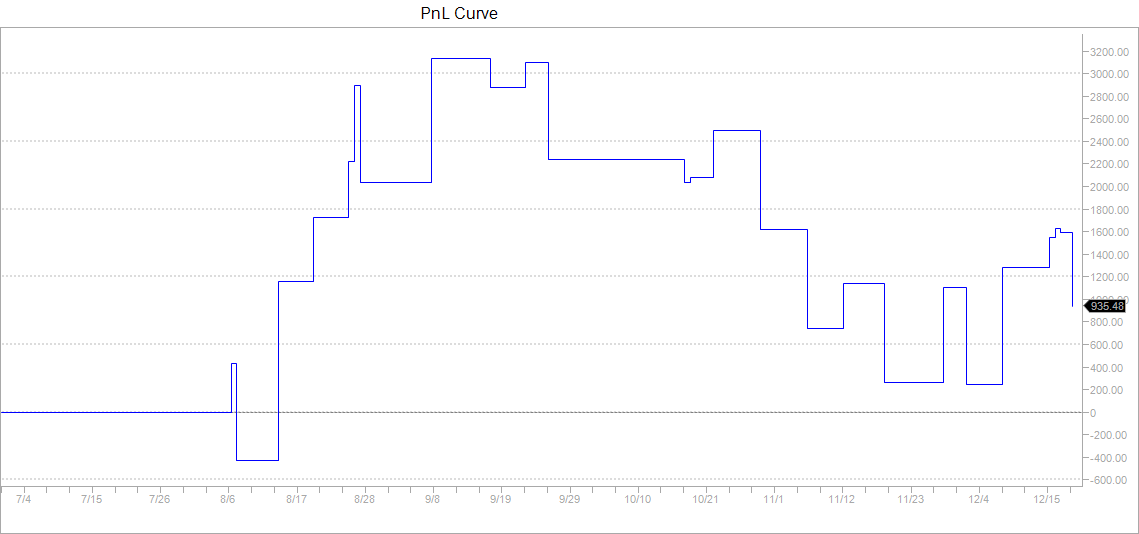}
\caption{Moving Average Crossover algorithm on test data set}
\label{fig:ma_test}
\end{figure}

\section{Conclusion}
In this paper, we presented a new method for learning Dynamic Bayesian Networks (DBN) using a new scoring metric that tackles the final usage of our DBN. The main purpose of this work is to present a new method for learning model parameters in DBNs that tackles the final cost function rather than EM that forces the model distribution to fit data at all cost and may result in poor final cost objective function. Thanks to evolutionary optimization techniques, we are able to find local optimum in polynomial time. Using information geometry, we show that the CMA ES method is theoretically sound and robust as it relies on the natural gradient induced by the Fisher information matrix. We conclude than possible extensions are to examine other black box optimization method to check their overall performance and to experiment this approach on other domains.

\newpage
\bibliography{mybib}

\end{document}